\documentclass{article}

\usepackage{microtype}
\usepackage{graphicx}
\usepackage{subfigure}
\usepackage{booktabs} % for professional tables

\usepackage{hyperref}
\usepackage{url}
\usepackage{graphicx}
\usepackage{algpseudocode}
\usepackage{graphicx}
\usepackage{amsmath}
\usepackage{amssymb}
\usepackage{mathtools}
\usepackage{amsthm}
\usepackage{caption}
\usepackage{adjustbox}
\usepackage{xcolor}
\usepackage{colortbl}
\usepackage{wrapfig}
\usepackage{subcaption}
\usepackage{float}
\usepackage{xcolor}
\usepackage{colortbl}
\definecolor{mycolor_gray}{HTML}{ECECEC}

\usepackage[accepted]{icml2025}

\usepackage{amsmath}
\usepackage{amssymb}
\usepackage{mathtools}
\usepackage{amsthm}

\usepackage[capitalize,noabbrev]{cleveref}

\theoremstyle{plain}

\theoremstyle{definition}

\theoremstyle{remark}

\usepackage[textsize=tiny]{todonotes}

\icmltitlerunning{HermesFlow: Seamlessly Closing the Gap in Multimodal Understanding and Generation}

\begin{document}

\twocolumn[
% \icmltitle{HermesFlow: Seamlessly Merging Multimodal Understanding and Generation}
\icmltitle{HermesFlow: \\Seamlessly Closing the Gap in Multimodal Understanding and Generation}
% \icmltitle{Reciprocal Optimization for Unified Multimodal Understanding and Generation}
% It is OKAY to include author information, even for blind
% submissions: the style file will automatically remove it for you
% unless you've provided the [accepted] option to the icml2025
% package.

% List of affiliations: The first argument should be a (short)
% identifier you will use later to specify author affiliations
% Academic affiliations should list Department, University, City, Region, Country
% Industry affiliations should list Company, City, Region, Country

% You can specify symbols, otherwise they are numbered in order.
% Ideally, you should not use this facility. Affiliations will be numbered
% in order of appearance and this is the preferred way.
\icmlsetsymbol{equal}{*}

\begin{icmlauthorlist}
\icmlauthor{Ling Yang}{equal,1}
\icmlauthor{Xinchen Zhang}{equal,2}\\
\icmlauthor{Ye Tian}{1}
\icmlauthor{Chenming Shang}{2}
\icmlauthor{Minghao Xu}{1,3}
\icmlauthor{Wentao Zhang}{1}
\icmlauthor{Bin Cui}{1}\\
\url{https://github.com/Gen-Verse/HermesFlow}

\end{icmlauthorlist}

\icmlaffiliation{1}{Peking University}
\icmlaffiliation{2}{Tsinghua University}
\icmlaffiliation{3}{Mila - Québec AI Institute}

\icmlcorrespondingauthor{Ling Yang}{yangling0818@163.com}
% \icmlcorrespondingauthor{Firstname2 Lastname2}{first2.last2@www.uk}

% You may provide any keywords that you
% find helpful for describing your paper; these are used to populate
% the "keywords" metadata in the PDF but will not be shown in the document
\icmlkeywords{Machine Learning, ICML}

\vskip 0.3in
]
% this must go after the closing bracket ] following \twocolumn[ ...

% This command actually creates the footnote in the first column
% listing the affiliations and the copyright notice.
% The command takes one argument, which is text to display at the start of the footnote.
% The \icmlEqualContribution command is standard text for equal contribution.
% Remove it (just {}) if you do not need this facility.

%\printAffiliationsAndNotice{}  % leave blank if no need to mention equal contribution
\printAffiliationsAndNotice{\icmlEqualContribution} % otherwise use the standard text.

\begin{abstract}
The remarkable success of the autoregressive paradigm has made significant advancement in Multimodal Large Language Models (MLLMs), with powerful models like Show-o, Transfusion and Emu3 achieving notable progress in unified image understanding and generation. For the first time, we uncover a common phenomenon: the understanding capabilities of MLLMs are typically stronger than their generative capabilities, with a significant gap between the two. Building on this insight, we propose \textbf{HermesFlow}, a simple yet general framework designed to seamlessly bridge the gap between understanding and generation in MLLMs. Specifically, we take the homologous data as input to curate homologous preference data of both understanding and generation. Through Pair-DPO and self-play iterative optimization, HermesFlow effectively aligns multimodal understanding and generation using homologous preference data. Extensive experiments demonstrate the significant superiority of our approach over prior methods, particularly in narrowing the gap between multimodal understanding and generation. These findings highlight the potential of HermesFlow as a general alignment framework for next-generation multimodal foundation models.
\end{abstract}

\section{Introduction}

The rapid advancement of Large Language Models (LLMs) \citep{o1,guo2025deepseek,BoT,yang2025reasonflux} has driven significant development in both multimodal understanding \citep{liu2024visual, zhu2023minigpt, li2023blip} and autoregressive image generation \citep{sun2024autoregressive, tian2024visual, fan2024fluid}. Recent studies \citep{team2024chameleon, li2024synergen, wu2024liquid, wu2024janus, ma2024janusflow} have focused on developing unified system capable of handling both multimodal understanding and generation. Powerful Multimodal Large Language Models (MLLMs) like Show-o \citep{xie2024show}, Transfusion \citep{zhou2024transfusion}, and Emu3 \citep{wang2024emu3}, employ a single transformer to unify these tasks, demonstrating remarkable performance across both domains.

\begin{figure}[t]
\vskip 0.3in
\begin{center}
\centerline{\includegraphics[width=\columnwidth]{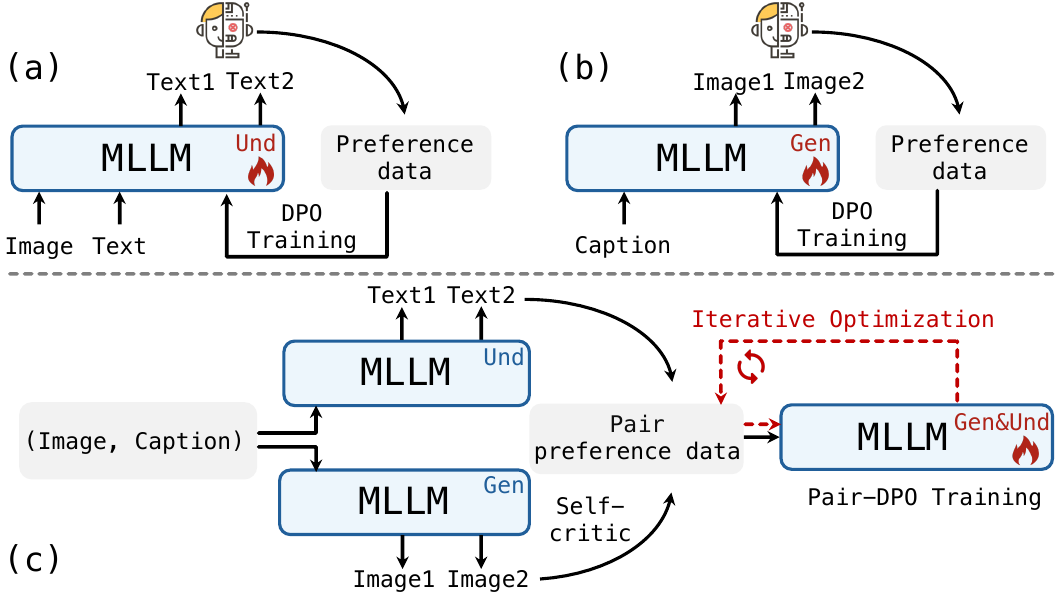}}
\caption{Architecture comparison between (a) DPO training improve multimodal understanding \citep{zhou2024calibrated, he2024self}, (b) DPO training improve multimodal generation \citep{wang2024emu3} and (c) our HermesFlow.}
\label{fig:compare}
\end{center}
\vskip -0.3in
\end{figure}

Recently, there has been growing interest in exploring the synergy between multimodal understanding and generation \citep{wu2024liquid, tong2024metamorph, dong2023dreamllm}. Liquid \citep{wu2024liquid} demonstrates that these two tasks are mutually beneficial: expanding the data for either understanding or generation enhances the performance of the other. Furthermore, MetaMorph \citep{tong2024metamorph} reveals that understanding data is more effective than generation data in improving both understanding and generation performance. However, these works jointly improve the understanding and generation capabilities of MLLMs from a data-level perspective but fail to consider the gap between them. It remains unclear whether a capability difference exists between them.

% However, these studies primarily examine the relationship between understanding and generation from a data-level perspective. It remains unclear whether a specific relationship exists between understanding and generation capabilities in MLLMs.

\begin{figure*}[t]
\vskip 0.3in
\begin{center}
\centerline{\includegraphics[width=1\textwidth]{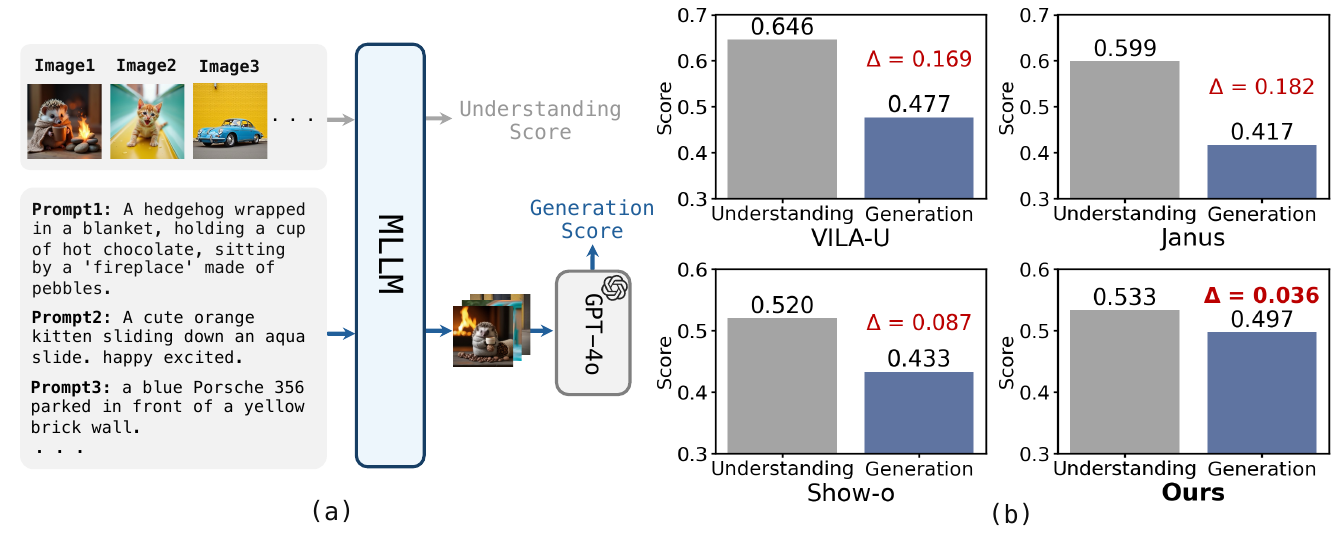}}
\vspace{-0.2in}
\caption{\textbf{Motivation of HermesFlow.} (a) A general pipeline to quantitatively assess the MLLM's performance of multimodal understanding and generation. (b) The imbalance between understanding and generation capabilities is a common phenomenon in MLLMs, and our method ssignificantly narrows this disparity. For detailed descriptions, please refer to \cref{gapscore}. }
\label{fig:motivation}
\end{center}
\vskip -0.2in
\end{figure*}

To quantitatively assess the performance of multimodal understanding and generation, we design a general pipeline, as illustrated in \cref{fig:motivation} (a). For any pretrained MLLM, input consists of (image, prompt/caption) pairs. For understanding tasks, MLLM is presented with multiple questions related to each image, and the final understanding score is calculated as the average accuracy of its answers. MLLM generates an image for each prompt, and these images are evaluated by posing the same set of questions using GPT-4o \citep{hurst2024gpt}, with the final generation score calculated based on the average accuracy of GPT-4o's answers. We employed this pipeline to evaluate multiple MLLMs. As shown in \cref{fig:motivation} (b), models like VILA-U \citep{wu2024vila}, Janus \citep{wu2024janus} and Show-o \citep{xie2024show} exhibit notably stronger understanding capabilities compared to their generation capabilities. Our experiments highlight a recurring phenomenon: \textbf{MLLMs consistently demonstrate superior understanding abilities over generation abilities, with a significant gap between them}.

% Direct Preference Optimization (DPO) \citep{rafailov2024direct} enables further finetuning of pretrained models using diverse preference data.

In the pretraining of MLLMs, simply increasing the training data for understanding or generation does not yield proportional improvements in both aspects \citep{tong2024metamorph}, leaving a significant gap between their understanding and generation capabilities. To bridge the gap between understanding and generation in MLLMs, we propose \textbf{\textit{HermesFlow}}, which collects paired understanding and generation preferences from homologous input data, and then employ a novel Pair-DPO post-training framework to seamlessly bridge the gap through the paired preference data. To curate understanding preference data, we enable MLLM to generate multiple captions for a single input image and filter paired understanding preference data using BERT similarity scores. To curate generation preference data, we prompt MLLM to generate multiple images from a single prompt and employ a self-critic-like approach to evaluate the images through self-VQA scoring, thereby filtering and selecting the paired generation preference data. Finally, we design Pair-DPO for preference alignment of homologous paired data, and through iterative optimization to simultaneously and progressively reduce the gap between understanding and generation following the same approach. We achieve the self-improvement of both understanding and generation of MLLM without incorporating any external high-quality training data.

We compare HermesFlow with previous work in \cref{fig:compare} and summarize our main contributions as follows:
\begin{itemize}
    \item An insightful discovery regarding a significant gap between the understanding and generation abilities of MLLMs, with understanding consistently outperforming generation.
    \item We propose a general multimodal self-improvement framework, \textit{HermesFlow}, using Pair-DPO based on homologous data to seamlessly close the gap between multimodal understanding and generation.
    \item Self-play iterative optimization paradigm is highly compatible with the multi-round enhancement of MLLMs. HermesFlow has potential as a general alignment framework for next-generation multimodal foundation models.
    \item Extensive qualitative and quantitative comparisons with previous powerful methods, such as Show-o, Janus and VILA-U, demonstrate the effectiveness and superiority of our method.
\end{itemize}

\section{Related Work}
\subsection{Unified Multimodal Understanding and Generation}
In recent years, a growing number of studies \citep{dong2023dreamllm, ge2024seed, wu2023next, ye2024x, ma2024janusflow, shi2024llamafusion} have explored unified multimodal models capable of both visual understanding and generation. Early methods \citep{dong2023dreamllm, tong2024metamorph, ge2024seed, sun2024generative, zhuang2024towards,zhang2024realcompo} leveraged diffusion models as external tools, where MLLMs generate conditions for visual generation \citep{yang2024mastering,tian2024videotetris} without having direct generative capabilities. For instance, DreamLLM \citep{dong2023dreamllm} introduces learnable embeddings called dream queries, which encapsulate the semantics encoded by MLLMs and serve as conditions for the diffusion decoder. More recently, inspired by the success of autoregressive paradigms, many studies \citep{team2024chameleon, xie2024show, zhou2024transfusion, qu2024tokenflow, xie2024muse, zhang2024fate, wang2024emu3} have shifted focus to representing and generating images using discrete visual tokens within a single transformer framework. For instance, Emu3 \citep{wang2024emu3} is trained solely with next-token prediction on a mixture of multimodal sequences using a single transformer. Janus \citep{wu2024janus} separates visual encoding into distinct pathways for multimodal understanding and generation while maintaining a unified transformer architecture. However, no existing research has focused on the relationship between the strengths of understanding and generation capabilities in MLLMs, which is essential for the balanced and sustainable development of these models.

\subsection{DPO in Multimodal LLMs}
Direct Preference Optimization (DPO) \citep{rafailov2024direct,zhang2024itercomp,yang2025supercorrect,yang2025reasonflux} enhances the performance of multimodal LLMs through the post-training process. In \cref{fig:compare}, we categorize these approaches into three types. Some methods \citep{zhou2024aligning, zhou2024calibrated, he2024self, zhang2024critic} utilize DPO to enhance understanding capability, as shown in \cref{fig:compare} (a). For instance, CSR \citep{zhou2024calibrated} enables the model to self-improve by iteratively generating candidate responses, evaluating the reward for each response, and curating preference data for finetuning. Other methods \citep{wang2024emu3} improve the generation capability of MLLMs through DPO as illustrated in \cref{fig:compare} (b). Emu3 \citep{wang2024emu3} generates a data pool and constructs a preference dataset through manual ranking, which is then used to optimize the model's generation capabilities via DPO. However, these models focus exclusively on enhancing either understanding or generation capabilities. In contrast, our approach uses Pair-DPO to effectively narrow the gap between the two, achieving mutual improvement.

\section{Preliminary}
\subsection{Next Token Prediction}

Next token prediction is a fundamental task in sequence modeling, where the goal is to estimate the conditional probability of the next token $x_t$ given its preceding context $x_{<t} = \{x_1, x_2, \dots, x_{t-1}\}$. Formally, for a sequence $\mathbf{x} = \{x_1, x_2, \dots, x_T\}$, the joint probability is factorized as:
\begin{equation}
    P(\mathbf{x}) =\! \prod_{t=1}^T P(x_t | x_1, x_2, \dots, x_{t-1})=\prod_{t=1}^T P(x_t | x_{<t})
\end{equation}
This factorization relies on the autoregressive assumption, where each token depends solely on its preceding tokens. During training, the model is optimized by minimizing the negative log-likelihood loss over the dataset:
\begin{equation}
    \mathcal{L} = - \frac{1}{T} \sum_{t=1}^T \log P(x_t | x_{<t})
\end{equation}
In autoregressive models, next-token prediction facilitates sequential generation by iteratively sampling tokens from the learned distribution $P(x_t | x_{<t})$. This approach is widely applicable multimodal domains such as visual understanding and visual generation.

\subsection{Direct Preference Optimization}

Direct Preference Optimization (DPO) provides a straightforward and efficient method by directly utilizing pairwise preference data to optimize the policy model. Specifically, given an input prompt $x$, and a preference data pair $(y_w, y_l)$, DPO aims to maximize the probability of the preferred output $y_w$ and minimize that of the undesirable output $y_l$. The optimization objective is formulated as:
\begin{equation}
\begin{aligned}
\mathcal{L}_{\text{DPO}}(\theta) = &-\mathbb{E}_{(x, y_w, y_l) \sim \mathcal{D}} \Bigg[ \log \sigma \\
&\Bigg( 
 \beta \log \frac{\pi_{\theta}(y_w \!\mid\! x)}{\pi_{\text{ref}}(y_w\! \mid \!x)} \!-\! \beta \log \frac{\pi_{\theta}(y_l \!\mid \!x)}{\pi_{\text{ref}}(y_l \!\mid \!x)} 
\Bigg) \Bigg]
\end{aligned}
\end{equation}

where  $\mathcal{D}$ is the pair-wise preference dataset, $\sigma$ is the sigmoid function, $\pi_{\theta}(\cdot\! \mid \!x)$ is the policy model to be optimized, $\pi_{\text{ref}}(\cdot \!\mid \!x)$  is the reference model kept unchanged during training, and the hyperparameter $\beta$ controls the distance from the reference model.

\begin{figure*}[t]
\vskip 0.4in
\begin{center}
\centerline{\includegraphics[width=1\textwidth]{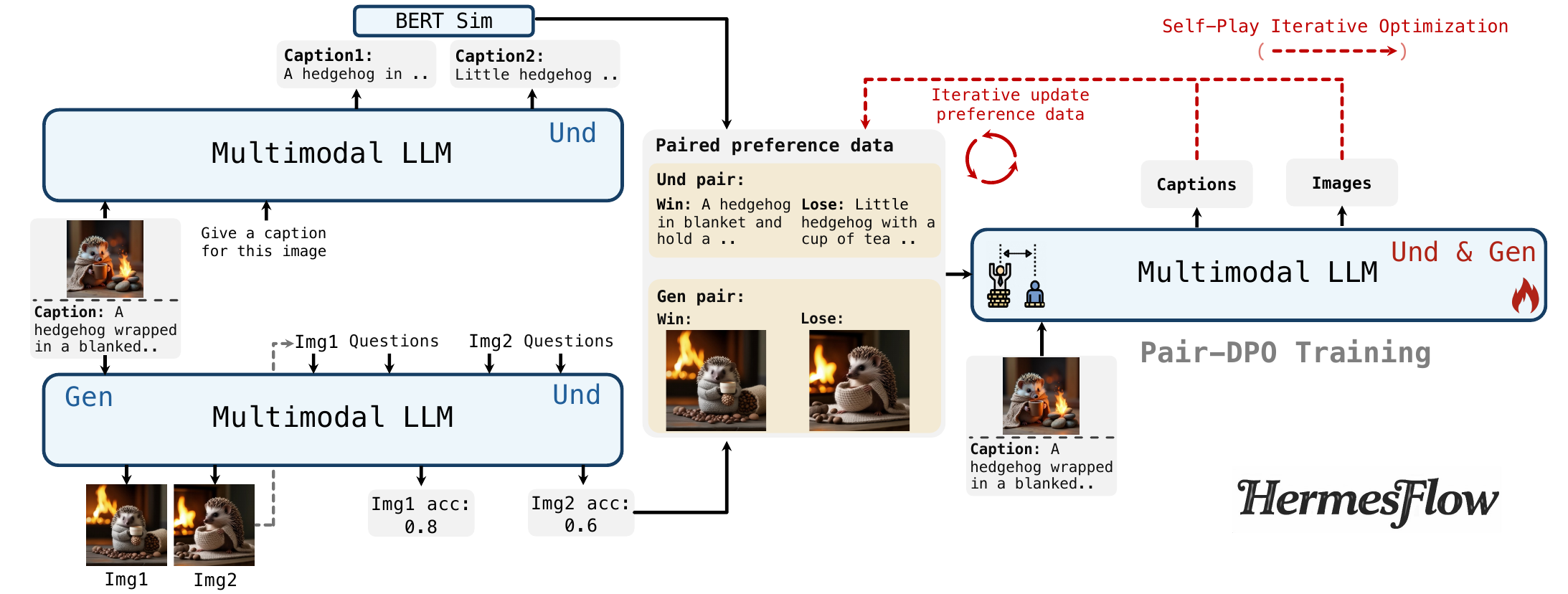}}
\caption{\textbf{Pipeline of HermesFlow.} We begin by curating paired data that captures both understanding and generation preferences from homologous input data. Leveraging this homologous preference data, we design Pair-DPO and employ self-play iterative optimization to seamlessly bridge the gap between multimodal understanding and generation. }
\label{pipeline}
\end{center}
\vskip -0.1in
\end{figure*}

\section{Method}
In this section, we present our method, HermesFlow, which curates pairwise preference data for both multimodal understanding and generation using homologous images and prompts, and seamlessly bridging the gap of multimodal understanding and generation through Pair-DPO training. An overview of HermesFlow is illustrated in \cref{pipeline}. In \cref{homopre}, we detail the methods for curating homologous preference data for multimodal understanding and generation, respectively. In \cref{pairdpo}, we propose the Pair-DPO training strategy to bridge the gap between multimodal understanding and generation. In \cref{iterative}, we introduce self-play iterative optimization, enabling the self-improvement of MLLM over multiple iterations.
\subsection{Curating Homologous Preference Data}
\label{homopre}
\paragraph{Homologous Input Data} The curation of both multimodal understanding and generation preference data begins with homologous data $(x, y)$, where $y$ represents the caption or prompt of the image $x$.

\paragraph{Understanding Preference Data} We focus on the image captioning task to collect understanding preference data, which reflects the ability of MLLMs to capture visual features, including object attributes, spatial relationships, and detailed elements of both the subject and background. Give an image $x$, a pretrained MLLM is used to generate $n$ different captions according to the input prompt: \textit{"Give a caption for this image."}. We then calculate the BERT similarity scores \cite{devlin2018bert} $s(y, x)$ between the original prompt $y$ and each of the $n$ captions. The caption with the highest BERT similarity score is selected as the winning sample $y_w$, while the one with the lowest score is chosen as the losing sample $y_l$. Following this process, we construct the pairwise understanding preference data.

\paragraph{Generation Preference Data} Starting with the caption or prompt $y$, we use the pretrained MLLM to randomly generate $n$ images. Given that MLLM's understanding abilities surpass its generation capabilities, we apply a self-critique or self-selection method for choosing the generated data.

Specifically, given the prompt $y$, we use TIFA \citep{hu2023tifa} to generate $q$ visual question-answer pairs, denoted as $\{(Q_1, A_1), (Q_2, A_2),\dots,(Q_q, A_q)\}$. For each generated image, we evaluate them based on the accuracy of the VQA responses provided by the MLLM:
\begin{equation}
    Acc(x_j) = \frac{1}{q} \sum_{i=1}^{q} \mathbb{I}(R_{j,i} = A_i),\ \  \forall j = 1, 2, \ldots, n \\
\end{equation}
\begin{equation}
    R_{j,i}=\text{MLLM}(x_j, Q_{j, i})
\end{equation}
where $ R_{j,i}$ represents the response of MLLM according to the input of image $x_j$ and question $Q_{j, i}$. We select the image with the highest accuracy as the winning sample $x_w$ and the one with the lowest accuracy as the losing sample $x_l$, while also ensuring that the highest accuracy exceeds 0.6. Using this process, we construct the pairwise generation preference data.

\paragraph{Homologous Output Preference Data} After curating understanding and generation preference data from homologous input $(x, y)$ as mentioned above,  where $y$ represents the caption or prompt of the image $x$, we obtain the homologous output preference data $\mathcal{D}$, denoted as $(x, y, x_w, x_l, y_w, y_l)$.

\subsection{Unified Enhancement with Pair-DPO}
\label{pairdpo}
Homologous preference paired data of understanding and generation indicate the optimized directions for both capabilities of a pretrained MLLM within the same semantic space. To achieve joint optimization and alignment of understanding and generation, we introduce Pair-DPO. The optimization objective of Pair-DPO can be formulated as:
\begin{equation}\label{pairdpo_equ}
    \mathcal{L}_{\text{Pair-DPO}}(\theta ) = 
-\mathbb{E}_{(x, y, x_w, x_l, y_w, y_l) \sim \mathcal{D}} 
\left[ 
    \log \sigma \left( 
        \Delta_{Und}\Delta_{Gen}
    \right) 
\right]
\end{equation}
\begin{equation}
    \Delta_{Und} = \beta \log \frac{\pi_\theta(y_w \mid x)}{\pi_{\text{ref}}(y_w \mid x)} 
        - 
        \beta \log \frac{\pi_\theta(y_l \mid x)}{\pi_{\text{ref}}(y_l \mid x)} 
\end{equation}
\begin{equation}
    \Delta_{Gen} = \beta \log \frac{\pi_\theta(x_w \mid y)}{\pi_{\text{ref}}(x_w \mid y)} - 
        \beta \log \frac{\pi_\theta(x_l \mid y)}{\pi_{\text{ref}}(x_l \mid y)} 
\end{equation}
where $\Delta_{Gen}$ and $\Delta_{Und}$ represent the preference differences in generation and understanding of MLLM, respectively. By using Pair-DPO to optimize homologous preference data jointly, we not only ensure mutual improvement in the understanding and generation capabilities of MLLM but also effectively narrow the gap between them. We provide the detailed derivation of the Pair-DPO optimization objective in \cref{derivation}.

\subsection{Self-Play Iterative Optimization}
\label{iterative}
To achieve comprehensive optimization and achieve a convergence gap in understanding and generation of MLLMs, we introduce a novel yet easy self-play iterative optimization using Pair-DPO with multiple turns.

Take understanding preference data as an example. We denote the preference data curated in round $i-1$ in \cref{homopre} as $(y^{i-1}_w, y^{i-1}_l)$. In the optimization of round $i$, the optimized MLLM generates $n$ new captions $(y^i_1, y^i_2, \dots, y^i_n)$ from the input of image $x$. The preference data is selected based on the following rules:
\begin{equation}
y^i_{\text{max}} = \arg\max_{k \in \{1,\ldots,n\}} \ s(y^i_k, y)
\end{equation}
\begin{equation}\label{eq:update_data}
(y^i_w, y^i_l) \!= \!
\begin{cases}\!
(y^i_{\text{max}}, y^{i-1}_w) & \!\text{if } \!\ s(y^i_{\text{max}}, y) \!>\! s(y^{i-1}_w, y) \\\!
(y^i_{\text{max}}, y^{i-1}_l) & \!\text{otherwise}
\end{cases}
\end{equation}
where $s(y_k^i, y)$ denotes the BERT similarity score between the generated caption $y_k^i$ and the homologous input caption $y$. Select the caption $y^i_{\text{max}}$ with the highest similarity score, which represents the local upper bound of the optimized MLLM’s understanding capability. If $s(y^i_{\text{max}}, y)> s(y^{i-1}_w, y)$, MLLM has effectively learned preference knowledge from the previous round. Therefore, it needs to be updated and further optimized using the higher-quality sample $y^i_{\text{max}}$ as the benchmark. Conversely, if $s(y^i_{\text{max}}, y)< s(y^{i-1}_w, y)$, effective optimization was not achieved in the previous round. In this case, it is necessary to update with simpler and clearer preference data $y^i_{\text{max}}$ as the winning sample to provide a smoother learning gradient. Through iterative optimization, we achieve self-improvement of MLLM without relying on any external high-quality training data.

\begin{algorithm}[t]
   \caption{The pseudocode of HermesFlow}
   \label{algorithm1}
    \begin{algorithmic}[1]
    \Statex {\bfseries Input:} Homologous data $(x, y)$, pretrained model $\text{MLLM}_\theta$ with parameters $\theta$
    \For {$i=0,\ldots ,iter$}
    \If {$i=0$}
        \State $y_w, y_l=\text{MLLM}_\theta^i(x)$ \ \ // Und preference data
        \State $x_w, x_l=\text{MLLM}_\theta^i(y)$ \ \ // Gen preference data
    \Else
        \State $y^i_1, y^i_2, \dots, y^i_n =\text{MLLM}_\theta^{i-1}(x)$
        \State $y^i_{\text{max}} = \arg\max_{k \in \{1,\ldots,n\}} \ s(y^i_k, x)$
        \State Update und-preference data using \cref{eq:update_data}
        \State $x^i_1, x^i_2, \dots, x^i_n =\text{MLLM}_\theta^{i-1}(y)$
        \State $x^i_{\text{max}} = \arg\max_{k \in \{1,\ldots,n\}} \ Acc(x^i_k)$
        \State Update gen-preference data using \cref{eq:update_data}
    \EndIf
    \State Optimize $\text{MLLM}_\theta^{i-1}$ to $\text{MLLM}_\theta^{i}$ using \cref{pairdpo_equ}
    \EndFor
        \end{algorithmic}
\end{algorithm}

\begin{table*}[t!]
\begin{center}{
\caption{Evaluation on multimodal understanding benchmarks. The baseline data is quoted from Show-o \citep{xie2024show}.}
\vspace{1em}
\label{tab:understanding}
\resizebox{.92\linewidth}{!}{ 
\begin{tabular}{l|ccccccc}
        \toprule
            Model & \# Params & POPE$\uparrow$ & MME$\uparrow$ & Flickr30k$\uparrow$ & VQAv2$_{\text{(test)}}$$\uparrow$ & GQA$\uparrow$ & MMMU$\uparrow$  \\
    	\midrule
            Gemini-Nano-1 \citep{team2023gemini} & 1.8B &- & - & - & 62.7 & - & 26.3 \\
            CoDI \citep{tang2024any}& - & - & - & 12.8 & - & - & - \\
            Emu \citep{sun2024generative}& 13B & - & - & 77.4 & 57.2 & - & - \\
            NExT-GPT \citep{wu2023next}& 13B & - & - & 84.5 & 66.7 & - & - \\
            SEED-X \citep{ge2024seed}& 17B & 84.2 & 1435.7 & 52.3 & - & 47.9 & 35.6 \\
            DreamLLM \citep{dong2023dreamllm}& 7B & - & - & - & 72.9 & - & - \\
            Chameleon \citep{team2024chameleon}&  34B & - & - & 74.7 & 66.0 & - & - \\
            Show-o \citep{xie2024show}& 1.3B & 80.0 & 1232.9 & 67.6 & 74.7 & 61.0 & 27.4 \\
            \textbf{HermesFlow (Ours)} & \cellcolor{mycolor_gray}{1.3B} & \cellcolor{mycolor_gray}{81.4} & \cellcolor{mycolor_gray}{1249.7} & \cellcolor{mycolor_gray}{69.2} &\cellcolor{mycolor_gray}{75.3}  &  \cellcolor{mycolor_gray}{61.7}&  \cellcolor{mycolor_gray}{28.3}\\      
        \bottomrule
\end{tabular}
}}
\end{center}
\vspace{-0.5mm}
\end{table*}

\begin{figure*}[t!]
\vskip 0.2in
\begin{center}
\centerline{\includegraphics[width=\textwidth]{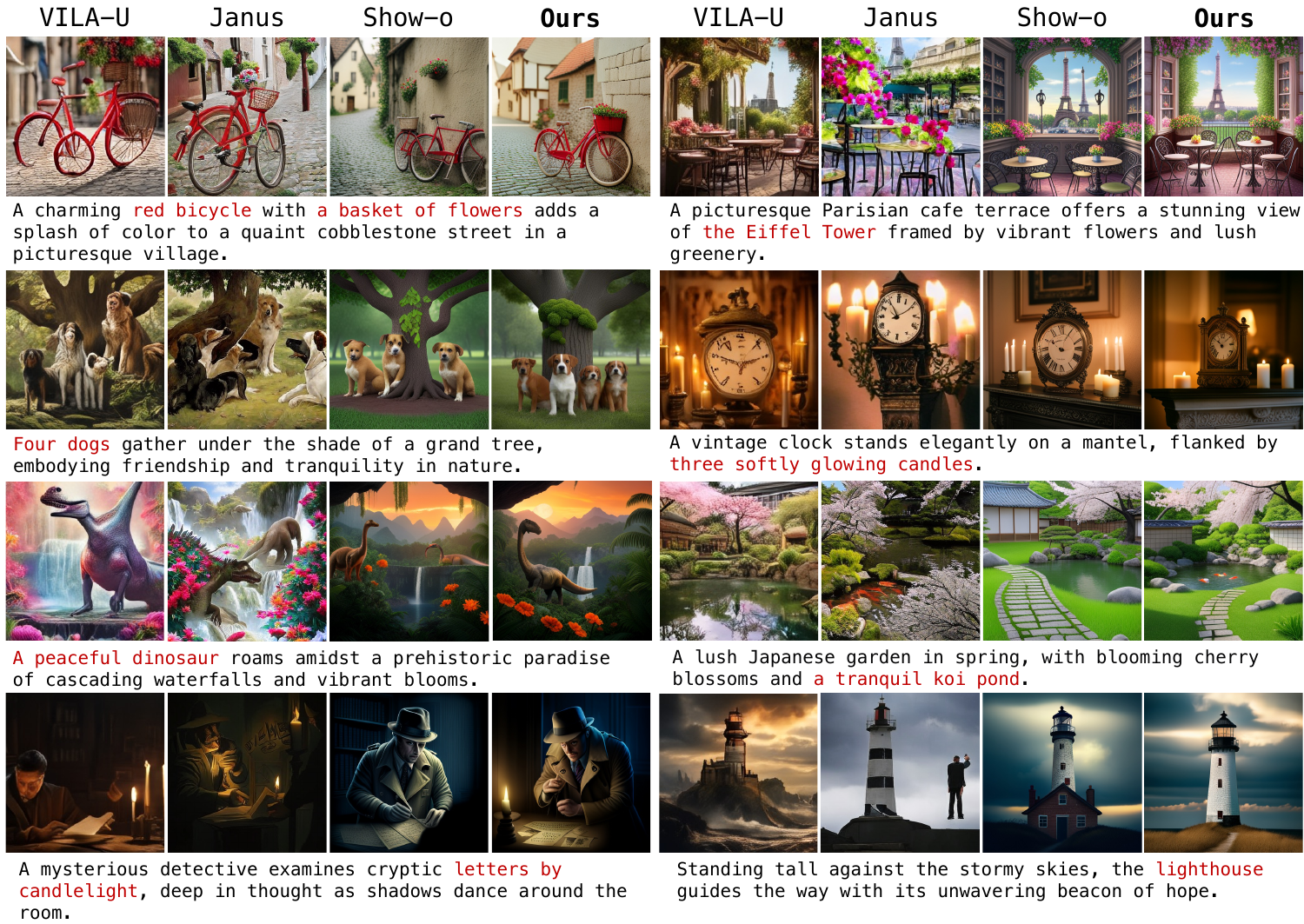}}
\caption{Qualitative comparison between our HermesFlow and three outstanding Multimodal LLMs VILA-U \citep{wu2024vila}, Janus \citep{wu2024janus}, and Show-o \citep{xie2024show}. Colored text denotes the advantages of HermesFlow in generated images.}
\label{fig:main_exp}
\end{center}
\vskip -0.2in
\end{figure*}

\begin{table*}[t!]
\begin{center}{
\caption{Evaluation on visual generation benchmarks: GenEval \citep{ghosh2024geneval} and DPG-Bench \citep{hu2024ella}. }
\vspace{0.5em}
\label{tab:generation}
\resizebox{1\textwidth}{!}{
\begin{tabular}{l l | c | c c c c c c  c | c}
\toprule
  & & & \multicolumn{7}{c|}{GenEval$\uparrow$} & \multicolumn{1}{c}{DPG-Bench$\uparrow$}\\
\multicolumn{2}{c|}{Methods}  & \#params  & Single Obj. & Two Obj. & Counting & Colors & Position & Color Attri. & Overall &Average \\
\midrule
\multicolumn{2}{l|}{\textit{Diffusion Model}} &&&&&&&&&\\
& LDM \citep{rombach2022high}  & 1.4B  & 0.92 & 0.29 & 0.23 & 0.70 & 0.02 & 0.05 & 0.37& -\\
& DALL-E 2 \citep{ramesh2022hierarchical} & 4.2B & 0.94 & 0.66 & 0.49 & 0.77 & 0.10 & 0.19 & 0.52 &-\\
& SD 1.5 \citep{rombach2022high}&860M & 0.94 & 0.37 & 0.27 & 0.72 & 0.05 & 0.07 & 0.40& 63.18\\
& SD 2.1 \citep{rombach2022high} &865M &  0.97 & 0.50 & 0.46 & 0.80 & 0.07 & 0.14 & 0.49& 68.09\\
\midrule
\multicolumn{2}{l|}{ \textit{Autoregressive Model}} &&&&&&&&& \\
& LlamaGen \citep{sun2024autoregressive}& 775M & 0.87 & 0.25 & 0.23 & 0.51 & 0.06 & 0.04 & 0.32& 65.16\\
& Emu \citep{sun2024generative}  & 14B  & 0.87 & 0.34 & 0.26& 0.56 & 0.07 & 0.06 & 0.36 & -\\
& Chameleon \citep{team2024chameleon}& 34B &0.89 &0.39  &0.28  & 0.66 &  0.08& 0.07  & 0.40 &-\\
& LWM \citep{liu2024world}& 7B & 0.93 & 0.41 & 0.46 & 0.79 & 0.09 & 0.15 & 0.47& -\\
& SEED-X \citep{ge2024seed}& 17B & 0.97 & 0.58 & 0.26 & 0.80 & 0.19 & 0.14 & 0.49& -\\
& Show-o \citep{xie2024show}& 1.3B & 0.95 & 0.52 & 0.49 & 0.82 & 0.11 & 0.28 & 0.53& 67.48\\
& Janus \citep{wu2024janus} & 1.3B & 0.97 & 0.68& 0.30&\textbf{0.84} &\textbf{0.46} &0.42 &0.61 & -\\
& \textbf{HermesFlow (Ours)} & \cellcolor{mycolor_gray}{1.3B}& \cellcolor{mycolor_gray}{\textbf{0.98}}& \cellcolor{mycolor_gray}{\textbf{0.84}}& \cellcolor{mycolor_gray}{\textbf{0.66}}& \cellcolor{mycolor_gray}{\underline{0.82}} &\cellcolor{mycolor_gray}{\underline{0.32}}  &\cellcolor{mycolor_gray}{\textbf{0.52}}  &\cellcolor{mycolor_gray}{\textbf{0.69}} &\cellcolor{mycolor_gray}{\textbf{70.22}}\\
\bottomrule
\end{tabular}
}
}
\end{center}
\vspace{-1em}
\end{table*}

\begin{table}[t]
\centering
\caption{MSCOCO zero-shot FID and CLIP-Score.}
\vspace{1em}
\label{tab:mscoco}
\resizebox{0.48\textwidth}{!}{ 
\begin{tabular}{l|cccc}
        \toprule
    	Method & \# Params  & FID$\downarrow$ & CLIP-Score$\uparrow$ \\
    	\midrule
            LDM \citep{rombach2022high}    & 1.4B   & 12.64& - \\
            DALL·E 2 \citep{ramesh2022hierarchical} & 6.5B   & 10.39& -\\
            SD 1.5 \citep{rombach2022high}  & 860M  & 9.62&  30.23\\
            SD 2.1 \citep{rombach2022high} & 865M & 8.03 & 30.87\\
            \midrule
            LlamaGen \citep{sun2024autoregressive} & 775M & 9.45 & 29.12\\
            Emu \citep{sun2024generative}& 14B & 11.02 &28.98 \\
            LWM \citep{liu2024world}& 7B  &12.68& - \\
            SEED-X \citep{ge2024seed}& 17B  &14.99&-  \\
            Show-o  \citep{xie2024show} & 1.3B  & 9.24& 30.63 \\
            \textbf{HermesFlow (Ours)} & \cellcolor{mycolor_gray}{1.3B}  & \cellcolor{mycolor_gray}{\textbf{9.07}}&  \cellcolor{mycolor_gray}{\textbf{31.08}}\\
        \bottomrule
\end{tabular}}
\vspace{-3mm}
\end{table} 

\section{Experiments}
\subsection{Experimental Setup}
\paragraph{Training Setup} We randomly select 5,000 image-caption pairs from JourneyDB \citep{sun2024journeydb} as our homologous input data. For the Visual Question Answering (VQA) data corresponding to each pair, we combine the VQA from JourneyDB with the VQA generated from TIFA \citep{hu2023tifa} for a comprehensive evaluation. Our HermesFlow is trained upon Showo \citep{xie2024show}, using a batch size of 4 for both caption and generation data over 3,000 steps. We employ the AdamW optimizer with a weight decay of 0.01, and an initial learning rate of 2e-5 with a cosine scheduling. The parameter $\beta$ for Pair-DPO is set to 0.2. All experiments are conducted under 8*NVIDIA A100 GPUs.

\paragraph{Evaluation Metrics} To assess multimodal understanding capabilities, we evaluate using POPE \citep{li2023evaluating}, MME \citep{fu2023mme}, Flickr30k \citep{plummer2015flickr30k}, VQAv2 \citep{goyal2017making}, GQA \citep{hudson2019gqa}, and MMMU \citep{yue2024mmmu}. For visual generation capabilities, we use GenEval \citep{ghosh2024geneval} and DPG-Bench \citep{hu2024ella} to evaluate the model's prompt-image alignment. We further assess image fidelity with FID \citep{heusel2017gans} and CLIP-Score \citep{radford2021learning}. Additionally, we conduct a comprehensive user study to objectively compare our model with other baselines.

\subsection{Main Results}

\paragraph{Multimodal Understanding Performances} \cref{tab:understanding} summarizes the comparison between our method and other leading MLLMs on multimodal understanding benchmarks. Notably, HermesFlow achieves similar or superior understanding performance compared to larger models like SEED-X and Chameleon, using less than 1/10 of the parameters. Additionally, HermesFlow demonstrates significant strengths across all metrics compared to Show-o, indicating that Pair-DPO effectively reduces the understanding-generation gap while maintaining or even enhancing understanding ability.

\paragraph{Image Generation Performances} As shown in \cref{fig:main_exp}, HermesFlow achieves superior generation results compared to three powerful Multimodal LLMs: VILA-U \citep{wu2024vila}, Janus \citep{wu2024janus}, and Show-o \citep{xie2024show}. Compared to its backbone, Show-o, HermesFlow demonstrates superior performance in generating object attributes and accurate counting. This improvement stems from its stronger understanding capabilities, which are utilized to filter generated images and achieve mutual refinement through Pair-DPO iteratively.

\begin{figure}[t]
\vskip 0.1in
\begin{center}
\centerline{\includegraphics[width=\columnwidth]{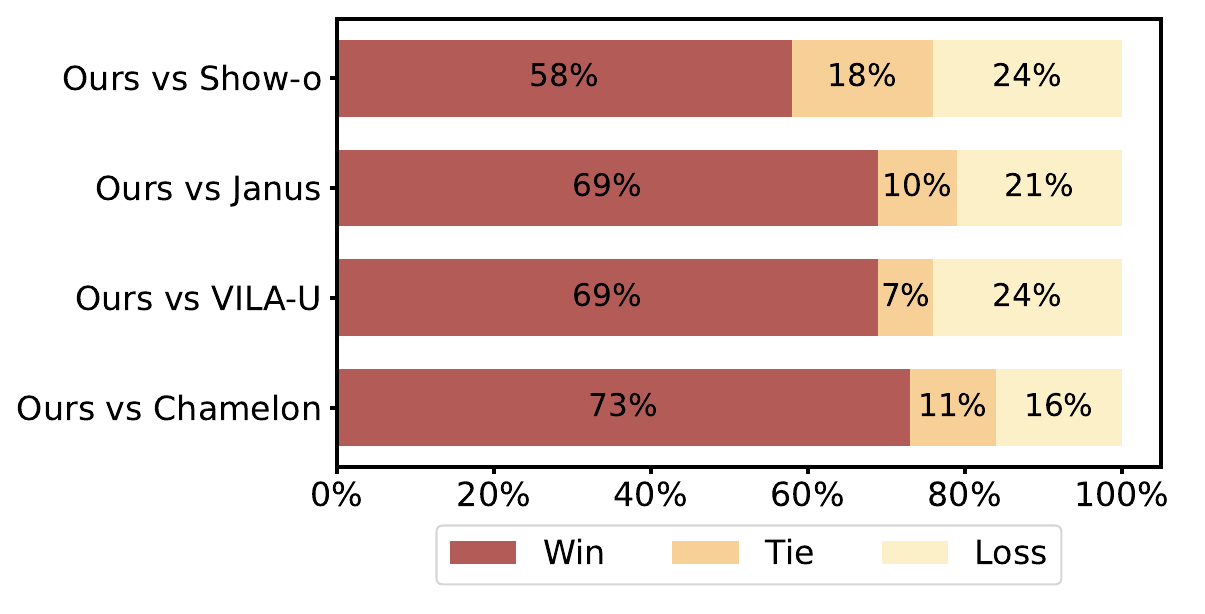}}
\vskip -0.1in
\caption{Results of user study.}
\label{fig:user_study}
\end{center}
\vskip -0.2in
\end{figure}

We compare HermesFlow with other visual generation models on GenEval \citep{ghosh2024geneval} and DPG-Bench \citep{hu2024ella}, as shown in \cref{tab:generation}. Compared to the diffusion-based generative model SD 2.1 \citep{rombach2022high}, HermesFlow demonstrates remarkable performance across all benchmarks. Furthermore, it surpasses larger autoregressive models, such as Chameleon \citep{team2024chameleon} and LWM \citep{liu2024world}. When compared to Show-o \citep{xie2024show}, HermesFlow exhibits significant strengths in object counting and positions, this is attributed to the critique of its superior understanding capability, which greatly enhances its visual generation performance in aspects such as object quantity, location, and attributes. We present the zero-shot FID \citep{heusel2017gans} and CLIP-Score \citep{radford2021learning} of HermesFlow on MSCOCO-30K in \cref{tab:mscoco}. The results clearly show that after the iterative optimization with Pair-DPO, HermesFlow achieves improved performance in both image fidelity and prompt-image alignment.

\begin{table*}[t!]
\centering
\caption{Quantitative assess of MLLM's Understanding and Generation Gap.}
\vspace{0.5em}
\label{tab:gap}
\resizebox{0.88\textwidth}{!}{ 
\begin{tabular}{l|ccccc}
        \toprule
    	Method & \# Params  & Understanding Score$\uparrow$ &Generation Score $\uparrow$ &Gap $\downarrow$\\
        \midrule
    	VILA-U \citep{wu2024vila}  \citep{xie2024show} & 7B  & 0.646& 0.477& 0.169\\
            Janus  \citep{wu2024janus} & 1.3B  & 0.599& 0.417& 0.182\\
            Show-o  \citep{xie2024show} & 1.3B  & 0.520& 0.433& 0.087\\
            \textbf{HermesFlow (Ours)} & \cellcolor{mycolor_gray}{1.3B}  & \cellcolor{mycolor_gray}{0.533}&  \cellcolor{mycolor_gray}{0.497}&\cellcolor{mycolor_gray}{\textbf{0.036}}\\
        \bottomrule
\end{tabular}}
\vspace{-3mm}
\end{table*} 
\paragraph{Quantitative assess of MLLM's Understanding and Generation Gap}
\label{gapscore}

\begin{table*}[t!]
\vspace{0.5em}
\begin{center}{
\caption{Comparison of Pair-DPO vs. DPO and the Effect of Pair-DPO Iterations.}
\vspace{0.5em}
\label{tab:pairdpo&dop}
\resizebox{.88\textwidth}{!}{
\begin{tabular}{l l | c c c | c c}
\toprule
  & &\multicolumn{3}{c|}{Understanding Bench} & \multicolumn{2}{c}{Generation Bench}\\
\multicolumn{2}{c|}{Methods}   & POPE$\uparrow$ & MME$\uparrow$ & MMMU$\uparrow$ & GenEval (Overall)$\uparrow$ & DPG-Bench (Average)$\uparrow$  \\
\midrule

& Show-o \citep{xie2024show} & 80.0 & 1232.9  & 27.4 & 0.53 & 67.48 \\
& DPO (Understanding)  & 80.8 & 1242.2& 27.8& 0.58&  67.88\\
& DPO (Generation)& 80.5  & 1239.3& 27.5&0.70 & 70.03 \\
\midrule
& Pair-DPO (Iter. 0) (Show-o) & 80.0 & 1232.9  & 27.4 & 0.53 & 67.48   \\
& Pair-DPO (Iter. 1) & 81.1 & 1246.7 & 28.0 & 0.68 & 70.19  \\
& Pair-DPO (Iter. 2)& 81.3 & 1248.3 & 28.1 & 0.69 & 70.21  \\
& Pair-DPO (Iter. 3)& \cellcolor{mycolor_gray}{\textbf{81.4}} & \cellcolor{mycolor_gray}{\textbf{1249.7}} & \cellcolor{mycolor_gray}{\textbf{28.3}} & \cellcolor{mycolor_gray}{\textbf{0.69}} & \cellcolor{mycolor_gray}{\textbf{70.22}}  \\

\bottomrule
\end{tabular}
}
}
\end{center}
\vspace{-0.5em}
\end{table*}

We also conducted a comprehensive user study to evaluate the effectiveness of HermesFlow in visual generation. As illustrated in \cref{fig:user_study}, we randomly selected 25 prompts for each comparison, and invited 35 users from diverse backgrounds to vote on image generation quality, collecting a total of 3,500 votes. Alignment between the generated images and the prompts was used as the primary evaluation criterion, with aesthetic quality and detail completeness considered under the same conditions. The results demonstrate that HermesFlow received widespread user approval in visual generation.

As shown in \cref{fig:motivation}, we use homologous data consisting of caption/prompt $y$ and image $x$ as input to evaluate the capability of understanding and generation respectively. The homologous data is randomly selected from JourneyDB \citep{sun2024journeydb}. For the understanding task, to ensure comprehensive and high-quality question-answer (QA) pairs, we first use TIFA \citep{hu2023tifa} to generate QA pairs based on the image and caption. These are then augmented with QA pairs from JourneyDB to create a more thorough and in-depth dataset. The final understanding score is calculated as the average accuracy of the answers. For the generation task, we use the prompt as input to generate an image for each prompt. These generated images are evaluated by posing the same set of questions to GPT-4o \citep{hurst2024gpt}, with the final generation score determined by the average accuracy of GPT-4o's answers. Since the generation capabilities of MLLMs are relatively limited, strict evaluation criteria are applied in cases of severe object blurring or significant loss of details. Therefore, GPT-4o is required to carefully analyze the completeness and authenticity of the objects involved in each question before providing answers. This evaluation pipeline was applied to multiple MLLMs, with the results presented in \cref{tab:gap}.

It is clear that a significant gap exists between multimodal understanding and generation in MLLM. HermesFlow seamlessly bridges this gap through self-play iterative optimization using Pair-DPO from homologous preference data.

\subsection{Ablation Study}
\paragraph{Pair-DPO vs. DPO}
Pair-DPO can simultaneously enhance both the understanding and generation capabilities of multimodal LLMs. As shown in \cref{tab:pairdpo&dop}, compared to DPO methods that rely solely on understanding or generation preference, a single round of Pair-DPO achieves superior performance by jointly optimizing both capabilities through the use of multimodal preference data. Furthermore, we observed that when using preference data from only one modality, whether understanding or generation, the capability of the other modality also improves. This demonstrates the same findings in MetaMorph \citep{tong2024metamorph} and Liquid \citep{wu2024liquid} that multimodal understanding and generation are synergistic.

\paragraph{Self-play Iterative Optimization} As shown in \cref{tab:pairdpo&dop}, we conducted an experimental analysis to examine the impact of iterations in self-play iterative optimization. It is evident that the first round of iterative optimization yields the most significant improvements in both understanding and generation. This is because the notable gap between the understanding and generation capabilities of MLLMs is most effectively bridged in the initial iteration. When the number of iterations exceeds 2, we observed that understanding ability continues to improve slightly, while generation ability remains almost stable. We argue that since generation is a fine-grained visual task, cross-capability transfer has limited impact on further enhancing generation ability in subsequent iterations. 

\paragraph{The Impact of Each Preference Sample Richness}

\begin{figure}[t]
\vskip 0.1in
\begin{center}
\centerline{\includegraphics[width=.5\textwidth]{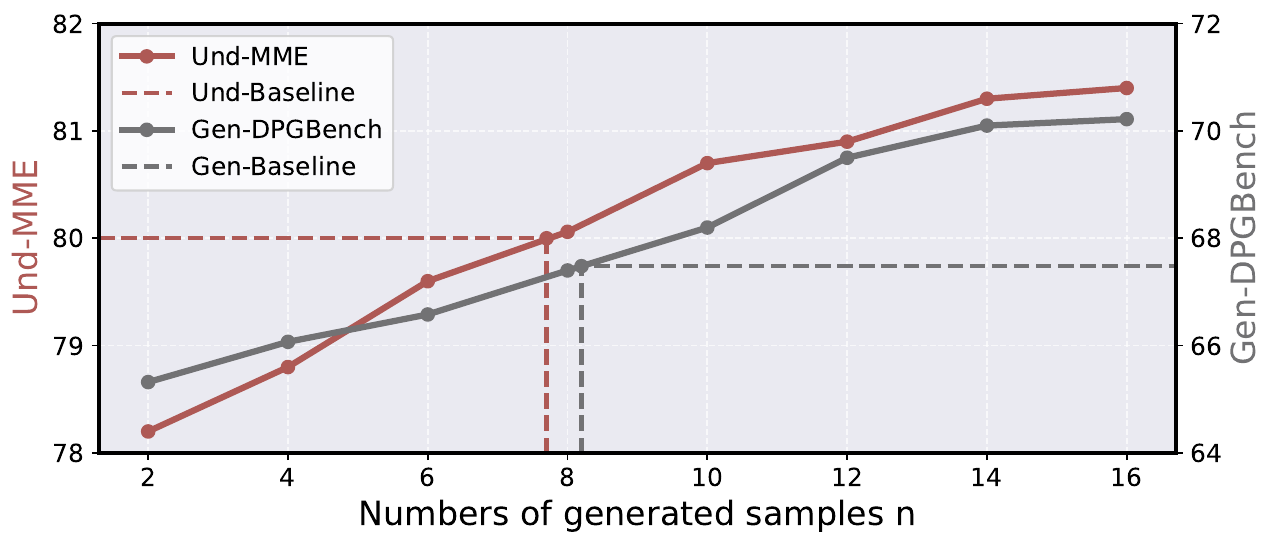}}
\caption{The influence of the richness of each preference sample.}
\label{fig:sample_size}
\end{center}
\vskip -0.3in
\end{figure}

The performance of Pair-DPO is largely influenced by the number of generated samples $n$ for both understanding and generation data. We conducted experiments to analyze the impact of $n$ on both understanding and generation in MLLMs, with results shown in \cref{fig:sample_size}. The dashed lines in the figure represent the performance of the baseline model, Show-o \citep{xie2024show}. 

When $n$ is too small, the model's understanding and generation performance decline. This is due to the insufficient number of samples and the limited capability of the baseline model, which leads to a noisy preference dataset and significantly impacts the results. However, as the sample size increases, it enables more accurate identification of the model’s optimal local upper bound, which in turn facilitates the curation of higher-quality preference data, leading to noticeable improvements in the understanding and generation capabilities of MLLMs.

Furthermore, \cref{fig:sample_size} reveals that achieving performance comparable to the baseline in generation requires more sampling data that understanding. This indicates that the generation capabilities of MLLMs are more sensitive to noise in the preference data, highlighting a greater need for high-quality generation data.

\section{Conclusion}
In this paper, we present a new MLLM alignment paradigm, HermesFlow, to seamlessly bridge the gap between multimodal understanding and generation. By iterative optimized with Pair-DPO using homologous preference data, HermesFlow successfully Improve the capabilities of both multimodal understanding and generation while narrowing the gap between them. Our extensive empirical evaluations across diverse understanding and generation benchmarks demonstrate the effectiveness of HermesFlow. However, due to current limitations in the number and capabilities of open-source MLLMs, HermesFlow has not yet been optimized across a wider range of backbones. In the future, we plan to extend this framework to more models. HermesFlow has the potential as a general alignment framework for next-generation multimodal foundation models.

\bibliography{example_paper}

\begin{thebibliography}{59}
\providecommand{\natexlab}[1]{#1}
\providecommand{\url}[1]{\texttt{#1}}
\expandafter\ifx\csname urlstyle\endcsname\relax
  \providecommand{\doi}[1]{doi: #1}\else
  \providecommand{\doi}{doi: \begingroup \urlstyle{rm}\Url}\fi

\bibitem[Devlin(2018)]{devlin2018bert}
Devlin, J.
\newblock Bert: Pre-training of deep bidirectional transformers for language understanding.
\newblock \emph{arXiv preprint arXiv:1810.04805}, 2018.

\bibitem[Dong et~al.(2023)Dong, Han, Peng, Qi, Ge, Yang, Zhao, Sun, Zhou, Wei, et~al.]{dong2023dreamllm}
Dong, R., Han, C., Peng, Y., Qi, Z., Ge, Z., Yang, J., Zhao, L., Sun, J., Zhou, H., Wei, H., et~al.
\newblock Dreamllm: Synergistic multimodal comprehension and creation.
\newblock \emph{arXiv preprint arXiv:2309.11499}, 2023.

\bibitem[Fan et~al.(2024)Fan, Li, Qin, Li, Sun, Rubinstein, Sun, He, and Tian]{fan2024fluid}
Fan, L., Li, T., Qin, S., Li, Y., Sun, C., Rubinstein, M., Sun, D., He, K., and Tian, Y.
\newblock Fluid: Scaling autoregressive text-to-image generative models with continuous tokens.
\newblock \emph{arXiv preprint arXiv:2410.13863}, 2024.

\bibitem[Fu et~al.(2023)Fu, Chen, Shen, Qin, Zhang, Lin, Yang, Zheng, Li, Sun, et~al.]{fu2023mme}
Fu, C., Chen, P., Shen, Y., Qin, Y., Zhang, M., Lin, X., Yang, J., Zheng, X., Li, K., Sun, X., et~al.
\newblock Mme: A comprehensive evaluation benchmark for multimodal large language models.
\newblock \emph{arXiv preprint arXiv:2306.13394}, 2023.

\bibitem[Ge et~al.(2024)Ge, Zhao, Zhu, Ge, Yi, Song, Li, Ding, and Shan]{ge2024seed}
Ge, Y., Zhao, S., Zhu, J., Ge, Y., Yi, K., Song, L., Li, C., Ding, X., and Shan, Y.
\newblock Seed-x: Multimodal models with unified multi-granularity comprehension and generation.
\newblock \emph{arXiv preprint arXiv:2404.14396}, 2024.

\bibitem[Ghosh et~al.(2024)Ghosh, Hajishirzi, and Schmidt]{ghosh2024geneval}
Ghosh, D., Hajishirzi, H., and Schmidt, L.
\newblock Geneval: An object-focused framework for evaluating text-to-image alignment.
\newblock \emph{Advances in Neural Information Processing Systems}, 36, 2024.

\bibitem[Goyal et~al.(2017)Goyal, Khot, Summers-Stay, Batra, and Parikh]{goyal2017making}
Goyal, Y., Khot, T., Summers-Stay, D., Batra, D., and Parikh, D.
\newblock Making the v in vqa matter: Elevating the role of image understanding in visual question answering.
\newblock In \emph{Proceedings of the IEEE conference on computer vision and pattern recognition}, pp.\  6904--6913, 2017.

\bibitem[Guo et~al.(2025)Guo, Yang, Zhang, Song, Zhang, Xu, Zhu, Ma, Wang, Bi, et~al.]{guo2025deepseek}
Guo, D., Yang, D., Zhang, H., Song, J., Zhang, R., Xu, R., Zhu, Q., Ma, S., Wang, P., Bi, X., et~al.
\newblock Deepseek-r1: Incentivizing reasoning capability in llms via reinforcement learning.
\newblock \emph{arXiv preprint arXiv:2501.12948}, 2025.

\bibitem[He et~al.(2024)He, Lin, Wang, Fung, and Ji]{he2024self}
He, J., Lin, H., Wang, Q., Fung, Y., and Ji, H.
\newblock Self-correction is more than refinement: A learning framework for visual and language reasoning tasks.
\newblock \emph{arXiv preprint arXiv:2410.04055}, 2024.

\bibitem[Heusel et~al.(2017)Heusel, Ramsauer, Unterthiner, Nessler, and Hochreiter]{heusel2017gans}
Heusel, M., Ramsauer, H., Unterthiner, T., Nessler, B., and Hochreiter, S.
\newblock Gans trained by a two time-scale update rule converge to a local nash equilibrium.
\newblock \emph{Advances in neural information processing systems}, 30, 2017.

\bibitem[Hu et~al.(2024)Hu, Wang, Fang, Fu, Cheng, and Yu]{hu2024ella}
Hu, X., Wang, R., Fang, Y., Fu, B., Cheng, P., and Yu, G.
\newblock Ella: Equip diffusion models with llm for enhanced semantic alignment.
\newblock \emph{arXiv preprint arXiv:2403.05135}, 2024.

\bibitem[Hu et~al.(2023)Hu, Liu, Kasai, Wang, Ostendorf, Krishna, and Smith]{hu2023tifa}
Hu, Y., Liu, B., Kasai, J., Wang, Y., Ostendorf, M., Krishna, R., and Smith, N.~A.
\newblock Tifa: Accurate and interpretable text-to-image faithfulness evaluation with question answering.
\newblock In \emph{Proceedings of the IEEE/CVF International Conference on Computer Vision}, pp.\  20406--20417, 2023.

\bibitem[Hudson \& Manning(2019)Hudson and Manning]{hudson2019gqa}
Hudson, D.~A. and Manning, C.~D.
\newblock Gqa: A new dataset for real-world visual reasoning and compositional question answering.
\newblock In \emph{Proceedings of the IEEE/CVF conference on computer vision and pattern recognition}, pp.\  6700--6709, 2019.

\bibitem[Hurst et~al.(2024)Hurst, Lerer, Goucher, Perelman, Ramesh, Clark, Ostrow, Welihinda, Hayes, Radford, et~al.]{hurst2024gpt}
Hurst, A., Lerer, A., Goucher, A.~P., Perelman, A., Ramesh, A., Clark, A., Ostrow, A., Welihinda, A., Hayes, A., Radford, A., et~al.
\newblock Gpt-4o system card.
\newblock \emph{arXiv preprint arXiv:2410.21276}, 2024.

\bibitem[Li et~al.(2024)Li, Tian, Shao, Zhu, Wang, Zhu, Dou, Wang, Li, Lu, et~al.]{li2024synergen}
Li, H., Tian, C., Shao, J., Zhu, X., Wang, Z., Zhu, J., Dou, W., Wang, X., Li, H., Lu, L., et~al.
\newblock Synergen-vl: Towards synergistic image understanding and generation with vision experts and token folding.
\newblock \emph{arXiv preprint arXiv:2412.09604}, 2024.

\bibitem[Li et~al.(2023{\natexlab{a}})Li, Li, Savarese, and Hoi]{li2023blip}
Li, J., Li, D., Savarese, S., and Hoi, S.
\newblock Blip-2: Bootstrapping language-image pre-training with frozen image encoders and large language models.
\newblock In \emph{International conference on machine learning}, pp.\  19730--19742. PMLR, 2023{\natexlab{a}}.

\bibitem[Li et~al.(2023{\natexlab{b}})Li, Du, Zhou, Wang, Zhao, and Wen]{li2023evaluating}
Li, Y., Du, Y., Zhou, K., Wang, J., Zhao, W.~X., and Wen, J.-R.
\newblock Evaluating object hallucination in large vision-language models.
\newblock \emph{arXiv preprint arXiv:2305.10355}, 2023{\natexlab{b}}.

\bibitem[Liu et~al.(2024{\natexlab{a}})Liu, Li, Wu, and Lee]{liu2024visual}
Liu, H., Li, C., Wu, Q., and Lee, Y.~J.
\newblock Visual instruction tuning.
\newblock \emph{Advances in neural information processing systems}, 36, 2024{\natexlab{a}}.

\bibitem[Liu et~al.(2024{\natexlab{b}})Liu, Yan, Zaharia, and Abbeel]{liu2024world}
Liu, H., Yan, W., Zaharia, M., and Abbeel, P.
\newblock World model on million-length video and language with blockwise ringattention.
\newblock \emph{CoRR}, 2024{\natexlab{b}}.

\bibitem[Ma et~al.(2024)Ma, Liu, Chen, Liu, Wu, Wu, Pan, Xie, Zhang, Zhao, et~al.]{ma2024janusflow}
Ma, Y., Liu, X., Chen, X., Liu, W., Wu, C., Wu, Z., Pan, Z., Xie, Z., Zhang, H., Zhao, L., et~al.
\newblock Janusflow: Harmonizing autoregression and rectified flow for unified multimodal understanding and generation.
\newblock \emph{arXiv preprint arXiv:2411.07975}, 2024.

\bibitem[OpenAI(2024)]{o1}
OpenAI.
\newblock Openai o1 system card.
\newblock \emph{preprint}, 2024.

\bibitem[Plummer et~al.(2015)Plummer, Wang, Cervantes, Caicedo, Hockenmaier, and Lazebnik]{plummer2015flickr30k}
Plummer, B.~A., Wang, L., Cervantes, C.~M., Caicedo, J.~C., Hockenmaier, J., and Lazebnik, S.
\newblock Flickr30k entities: Collecting region-to-phrase correspondences for richer image-to-sentence models.
\newblock In \emph{Proceedings of the IEEE international conference on computer vision}, pp.\  2641--2649, 2015.

\bibitem[Qu et~al.(2024)Qu, Zhang, Liu, Wang, Jiang, Gao, Ye, Du, Yuan, and Wu]{qu2024tokenflow}
Qu, L., Zhang, H., Liu, Y., Wang, X., Jiang, Y., Gao, Y., Ye, H., Du, D.~K., Yuan, Z., and Wu, X.
\newblock Tokenflow: Unified image tokenizer for multimodal understanding and generation.
\newblock \emph{arXiv preprint arXiv:2412.03069}, 2024.

\bibitem[Radford et~al.(2021)Radford, Kim, Hallacy, Ramesh, Goh, Agarwal, Sastry, Askell, Mishkin, Clark, et~al.]{radford2021learning}
Radford, A., Kim, J.~W., Hallacy, C., Ramesh, A., Goh, G., Agarwal, S., Sastry, G., Askell, A., Mishkin, P., Clark, J., et~al.
\newblock Learning transferable visual models from natural language supervision.
\newblock In \emph{International conference on machine learning}, pp.\  8748--8763. PMLR, 2021.

\bibitem[Rafailov et~al.(2024)Rafailov, Sharma, Mitchell, Manning, Ermon, and Finn]{rafailov2024direct}
Rafailov, R., Sharma, A., Mitchell, E., Manning, C.~D., Ermon, S., and Finn, C.
\newblock Direct preference optimization: Your language model is secretly a reward model.
\newblock \emph{Advances in Neural Information Processing Systems}, 36, 2024.

\bibitem[Ramesh et~al.(2022)Ramesh, Dhariwal, Nichol, Chu, and Chen]{ramesh2022hierarchical}
Ramesh, A., Dhariwal, P., Nichol, A., Chu, C., and Chen, M.
\newblock Hierarchical text-conditional image generation with clip latents.
\newblock \emph{arXiv preprint arXiv:2204.06125}, 1\penalty0 (2):\penalty0 3, 2022.

\bibitem[Rombach et~al.(2022)Rombach, Blattmann, Lorenz, Esser, and Ommer]{rombach2022high}
Rombach, R., Blattmann, A., Lorenz, D., Esser, P., and Ommer, B.
\newblock High-resolution image synthesis with latent diffusion models.
\newblock In \emph{Proceedings of the IEEE/CVF conference on computer vision and pattern recognition}, pp.\  10684--10695, 2022.

\bibitem[Shi et~al.(2024)Shi, Han, Zhou, Liang, Lin, Zettlemoyer, and Yu]{shi2024llamafusion}
Shi, W., Han, X., Zhou, C., Liang, W., Lin, X.~V., Zettlemoyer, L., and Yu, L.
\newblock Llamafusion: Adapting pretrained language models for multimodal generation.
\newblock \emph{arXiv preprint arXiv:2412.15188}, 2024.

\bibitem[Sun et~al.(2024{\natexlab{a}})Sun, Pan, Ge, Li, Duan, Wu, Zhang, Zhou, Qin, Wang, et~al.]{sun2024journeydb}
Sun, K., Pan, J., Ge, Y., Li, H., Duan, H., Wu, X., Zhang, R., Zhou, A., Qin, Z., Wang, Y., et~al.
\newblock Journeydb: A benchmark for generative image understanding.
\newblock \emph{Advances in Neural Information Processing Systems}, 36, 2024{\natexlab{a}}.

\bibitem[Sun et~al.(2024{\natexlab{b}})Sun, Jiang, Chen, Zhang, Peng, Luo, and Yuan]{sun2024autoregressive}
Sun, P., Jiang, Y., Chen, S., Zhang, S., Peng, B., Luo, P., and Yuan, Z.
\newblock Autoregressive model beats diffusion: Llama for scalable image generation.
\newblock \emph{arXiv preprint arXiv:2406.06525}, 2024{\natexlab{b}}.

\bibitem[Sun et~al.(2024{\natexlab{c}})Sun, Cui, Zhang, Zhang, Yu, Wang, Rao, Liu, Huang, and Wang]{sun2024generative}
Sun, Q., Cui, Y., Zhang, X., Zhang, F., Yu, Q., Wang, Y., Rao, Y., Liu, J., Huang, T., and Wang, X.
\newblock Generative multimodal models are in-context learners.
\newblock In \emph{Proceedings of the IEEE/CVF Conference on Computer Vision and Pattern Recognition}, pp.\  14398--14409, 2024{\natexlab{c}}.

\bibitem[Tang et~al.(2024)Tang, Yang, Zhu, Zeng, and Bansal]{tang2024any}
Tang, Z., Yang, Z., Zhu, C., Zeng, M., and Bansal, M.
\newblock Any-to-any generation via composable diffusion.
\newblock \emph{Advances in Neural Information Processing Systems}, 36, 2024.

\bibitem[Team(2024)]{team2024chameleon}
Team, C.
\newblock Chameleon: Mixed-modal early-fusion foundation models.
\newblock \emph{arXiv preprint arXiv:2405.09818}, 2024.

\bibitem[Team et~al.(2023)Team, Anil, Borgeaud, Alayrac, Yu, Soricut, Schalkwyk, Dai, Hauth, Millican, et~al.]{team2023gemini}
Team, G., Anil, R., Borgeaud, S., Alayrac, J.-B., Yu, J., Soricut, R., Schalkwyk, J., Dai, A.~M., Hauth, A., Millican, K., et~al.
\newblock Gemini: a family of highly capable multimodal models.
\newblock \emph{arXiv preprint arXiv:2312.11805}, 2023.

\bibitem[Tian et~al.(2024{\natexlab{a}})Tian, Jiang, Yuan, Peng, and Wang]{tian2024visual}
Tian, K., Jiang, Y., Yuan, Z., Peng, B., and Wang, L.
\newblock Visual autoregressive modeling: Scalable image generation via next-scale prediction.
\newblock \emph{arXiv preprint arXiv:2404.02905}, 2024{\natexlab{a}}.

\bibitem[Tian et~al.(2024{\natexlab{b}})Tian, Yang, Yang, Gao, Deng, Chen, Wang, Yu, Tao, Wan, et~al.]{tian2024videotetris}
Tian, Y., Yang, L., Yang, H., Gao, Y., Deng, Y., Chen, J., Wang, X., Yu, Z., Tao, X., Wan, P., et~al.
\newblock Videotetris: Towards compositional text-to-video generation.
\newblock \emph{arXiv preprint arXiv:2406.04277}, 2024{\natexlab{b}}.

\bibitem[Tong et~al.(2024)Tong, Fan, Zhu, Xiong, Chen, Sinha, Rabbat, LeCun, Xie, and Liu]{tong2024metamorph}
Tong, S., Fan, D., Zhu, J., Xiong, Y., Chen, X., Sinha, K., Rabbat, M., LeCun, Y., Xie, S., and Liu, Z.
\newblock Metamorph: Multimodal understanding and generation via instruction tuning.
\newblock \emph{arXiv preprint arXiv:2412.14164}, 2024.

\bibitem[Wang et~al.(2024)Wang, Zhang, Luo, Sun, Cui, Wang, Zhang, Wang, Li, Yu, et~al.]{wang2024emu3}
Wang, X., Zhang, X., Luo, Z., Sun, Q., Cui, Y., Wang, J., Zhang, F., Wang, Y., Li, Z., Yu, Q., et~al.
\newblock Emu3: Next-token prediction is all you need.
\newblock \emph{arXiv preprint arXiv:2409.18869}, 2024.

\bibitem[Wu et~al.(2024{\natexlab{a}})Wu, Chen, Wu, Ma, Liu, Pan, Liu, Xie, Yu, Ruan, et~al.]{wu2024janus}
Wu, C., Chen, X., Wu, Z., Ma, Y., Liu, X., Pan, Z., Liu, W., Xie, Z., Yu, X., Ruan, C., et~al.
\newblock Janus: Decoupling visual encoding for unified multimodal understanding and generation.
\newblock \emph{arXiv preprint arXiv:2410.13848}, 2024{\natexlab{a}}.

\bibitem[Wu et~al.(2024{\natexlab{b}})Wu, Jiang, Ma, Liu, Zhao, Yuan, Bai, and Bai]{wu2024liquid}
Wu, J., Jiang, Y., Ma, C., Liu, Y., Zhao, H., Yuan, Z., Bai, S., and Bai, X.
\newblock Liquid: Language models are scalable multi-modal generators.
\newblock \emph{arXiv preprint arXiv:2412.04332}, 2024{\natexlab{b}}.

\bibitem[Wu et~al.(2023)Wu, Fei, Qu, Ji, and Chua]{wu2023next}
Wu, S., Fei, H., Qu, L., Ji, W., and Chua, T.-S.
\newblock Next-gpt: Any-to-any multimodal llm.
\newblock \emph{arXiv preprint arXiv:2309.05519}, 2023.

\bibitem[Wu et~al.(2024{\natexlab{c}})Wu, Zhang, Chen, Tang, Li, Fang, Zhu, Xie, Yin, Yi, et~al.]{wu2024vila}
Wu, Y., Zhang, Z., Chen, J., Tang, H., Li, D., Fang, Y., Zhu, L., Xie, E., Yin, H., Yi, L., et~al.
\newblock Vila-u: a unified foundation model integrating visual understanding and generation.
\newblock \emph{arXiv preprint arXiv:2409.04429}, 2024{\natexlab{c}}.

\bibitem[Xie et~al.(2024{\natexlab{a}})Xie, Mao, Bai, Zhang, Wang, Lin, Gu, Chen, Yang, and Shou]{xie2024show}
Xie, J., Mao, W., Bai, Z., Zhang, D.~J., Wang, W., Lin, K.~Q., Gu, Y., Chen, Z., Yang, Z., and Shou, M.~Z.
\newblock Show-o: One single transformer to unify multimodal understanding and generation.
\newblock \emph{arXiv preprint arXiv:2408.12528}, 2024{\natexlab{a}}.

\bibitem[Xie et~al.(2024{\natexlab{b}})Xie, Du, Song, and Liu]{xie2024muse}
Xie, R., Du, C., Song, P., and Liu, C.
\newblock Muse-vl: Modeling unified vlm through semantic discrete encoding.
\newblock \emph{arXiv preprint arXiv:2411.17762}, 2024{\natexlab{b}}.

\bibitem[Yang et~al.(2024{\natexlab{a}})Yang, Yu, Meng, Xu, Ermon, and Bin]{yang2024mastering}
Yang, L., Yu, Z., Meng, C., Xu, M., Ermon, S., and Bin, C.
\newblock Mastering text-to-image diffusion: Recaptioning, planning, and generating with multimodal llms.
\newblock In \emph{Forty-first International Conference on Machine Learning}, 2024{\natexlab{a}}.

\bibitem[Yang et~al.(2024{\natexlab{b}})Yang, Yu, Zhang, Cao, Xu, Zhang, Gonzalez, and Cui]{BoT}
Yang, L., Yu, Z., Zhang, T., Cao, S., Xu, M., Zhang, W., Gonzalez, J.~E., and Cui, B.
\newblock Buffer of thoughts: Thought-augmented reasoning with large language models.
\newblock \emph{Advances in Neural Information Processing Systems}, 2024{\natexlab{b}}.

\bibitem[Yang et~al.(2025{\natexlab{a}})Yang, Yu, Cui, and Wang]{yang2025reasonflux}
Yang, L., Yu, Z., Cui, B., and Wang, M.
\newblock Reasonflux: Hierarchical llm reasoning via scaling thought templates.
\newblock \emph{arXiv preprint arXiv:2502.06772}, 2025{\natexlab{a}}.

\bibitem[Yang et~al.(2025{\natexlab{b}})Yang, Yu, Zhang, Xu, Gonzalez, Cui, and Yan]{yang2025supercorrect}
Yang, L., Yu, Z., Zhang, T., Xu, M., Gonzalez, J.~E., Cui, B., and Yan, S.
\newblock Supercorrect: Supervising and correcting language models with error-driven insights.
\newblock In \emph{International Conference on Learning Representations}, 2025{\natexlab{b}}.

\bibitem[Ye et~al.(2024)Ye, Huang, Lu, Yu, Ping, Tao, Kautz, Han, Xu, Molchanov, et~al.]{ye2024x}
Ye, H., Huang, D.-A., Lu, Y., Yu, Z., Ping, W., Tao, A., Kautz, J., Han, S., Xu, D., Molchanov, P., et~al.
\newblock X-vila: Cross-modality alignment for large language model.
\newblock \emph{arXiv preprint arXiv:2405.19335}, 2024.

\bibitem[Yue et~al.(2024)Yue, Ni, Zhang, Zheng, Liu, Zhang, Stevens, Jiang, Ren, Sun, et~al.]{yue2024mmmu}
Yue, X., Ni, Y., Zhang, K., Zheng, T., Liu, R., Zhang, G., Stevens, S., Jiang, D., Ren, W., Sun, Y., et~al.
\newblock Mmmu: A massive multi-discipline multimodal understanding and reasoning benchmark for expert agi.
\newblock In \emph{Proceedings of the IEEE/CVF Conference on Computer Vision and Pattern Recognition}, pp.\  9556--9567, 2024.

\bibitem[Zhang et~al.(2024{\natexlab{a}})Zhang, Lei, Li, Wang, Liu, Yang, Li, Wang, Yang, Wu, et~al.]{zhang2024critic}
Zhang, D., Lei, J., Li, J., Wang, X., Liu, Y., Yang, Z., Li, J., Wang, W., Yang, S., Wu, J., et~al.
\newblock Critic-v: Vlm critics help catch vlm errors in multimodal reasoning.
\newblock \emph{arXiv preprint arXiv:2411.18203}, 2024{\natexlab{a}}.

\bibitem[Zhang et~al.(2024{\natexlab{b}})Zhang, Wu, Liang, Gong, Hu, Yao, Cao, and Zhu]{zhang2024fate}
Zhang, J., Wu, Z., Liang, Z., Gong, Y., Hu, D., Yao, Y., Cao, X., and Zhu, H.
\newblock Fate: Full-head gaussian avatar with textural editing from monocular video.
\newblock \emph{arXiv preprint arXiv:2411.15604}, 2024{\natexlab{b}}.

\bibitem[Zhang et~al.(2024{\natexlab{c}})Zhang, Yang, Cai, Yu, Wang, Tian, Xu, Tang, Yang, Bin, et~al.]{zhang2024realcompo}
Zhang, X., Yang, L., Cai, Y., Yu, Z., Wang, K.-N., Tian, Y., Xu, M., Tang, Y., Yang, Y., Bin, C., et~al.
\newblock Realcompo: Balancing realism and compositionality improves text-to-image diffusion models.
\newblock In \emph{The Thirty-eighth Annual Conference on Neural Information Processing Systems}, 2024{\natexlab{c}}.

\bibitem[Zhang et~al.(2024{\natexlab{d}})Zhang, Yang, Li, Cai, Xie, Tang, Yang, Wang, and Cui]{zhang2024itercomp}
Zhang, X., Yang, L., Li, G., Cai, Y., Xie, J., Tang, Y., Yang, Y., Wang, M., and Cui, B.
\newblock Itercomp: Iterative composition-aware feedback learning from model gallery for text-to-image generation.
\newblock \emph{arXiv preprint arXiv:2410.07171}, 2024{\natexlab{d}}.

\bibitem[Zhou et~al.(2024{\natexlab{a}})Zhou, Yu, Babu, Tirumala, Yasunaga, Shamis, Kahn, Ma, Zettlemoyer, and Levy]{zhou2024transfusion}
Zhou, C., Yu, L., Babu, A., Tirumala, K., Yasunaga, M., Shamis, L., Kahn, J., Ma, X., Zettlemoyer, L., and Levy, O.
\newblock Transfusion: Predict the next token and diffuse images with one multi-modal model.
\newblock \emph{arXiv preprint arXiv:2408.11039}, 2024{\natexlab{a}}.

\bibitem[Zhou et~al.(2024{\natexlab{b}})Zhou, Cui, Rafailov, Finn, and Yao]{zhou2024aligning}
Zhou, Y., Cui, C., Rafailov, R., Finn, C., and Yao, H.
\newblock Aligning modalities in vision large language models via preference fine-tuning.
\newblock \emph{arXiv preprint arXiv:2402.11411}, 2024{\natexlab{b}}.

\bibitem[Zhou et~al.(2024{\natexlab{c}})Zhou, Fan, Cheng, Yang, Chen, Cui, Wang, Li, Zhang, and Yao]{zhou2024calibrated}
Zhou, Y., Fan, Z., Cheng, D., Yang, S., Chen, Z., Cui, C., Wang, X., Li, Y., Zhang, L., and Yao, H.
\newblock Calibrated self-rewarding vision language models.
\newblock \emph{arXiv preprint arXiv:2405.14622}, 2024{\natexlab{c}}.

\bibitem[Zhu et~al.(2023)Zhu, Chen, Shen, Li, and Elhoseiny]{zhu2023minigpt}
Zhu, D., Chen, J., Shen, X., Li, X., and Elhoseiny, M.
\newblock Minigpt-4: Enhancing vision-language understanding with advanced large language models.
\newblock \emph{arXiv preprint arXiv:2304.10592}, 2023.

\bibitem[Zhuang et~al.(2024)Zhuang, He, Zhang, Wang, Zhu, Yao, Zhu, Cao, and Zhu]{zhuang2024towards}
Zhuang, Y., He, Y., Zhang, J., Wang, Y., Zhu, J., Yao, Y., Zhu, S., Cao, X., and Zhu, H.
\newblock Towards native generative model for 3d head avatar.
\newblock \emph{arXiv preprint arXiv:2410.01226}, 2024.

\end{thebibliography}
\bibliographystyle{icml2025}

%%%%%%%%%%%%%%%%%%%%%%%%%%%%%%%%%%%%%%%%%%%%%%%%%%%%%%%%%%%%%%%%%%%%%%%%%%%%%%%
%%%%%%%%%%%%%%%%%%%%%%%%%%%%%%%%%%%%%%%%%%%%%%%%%%%%%%%%%%%%%%%%%%%%%%%%%%%%%%%
% APPENDIX
%%%%%%%%%%%%%%%%%%%%%%%%%%%%%%%%%%%%%%%%%%%%%%%%%%%%%%%%%%%%%%%%%%%%%%%%%%%%%%%
%%%%%%%%%%%%%%%%%%%%%%%%%%%%%%%%%%%%%%%%%%%%%%%%%%%%%%%%%%%%%%%%%%%%%%%%%%%%%%%
\newpage
\appendix
\onecolumn

\section{Derivation of the Pair-DPO Optimization Objective}
\label{derivation}
Considering that the optimization objective of standard Direct Preference Optimization is:
\begin{equation}
\mathcal{L}_{\text{DPO}}(\theta) = -\mathbb{E}_{(x, y_w, y_l) \sim \mathcal{D}} \Bigg[ \log \sigma \Bigg( 
 \beta \log \frac{\pi_{\theta}(y_w \!\mid\! x)}{\pi_{\text{ref}}(y_w\! \mid \!x)} \!-\! \beta \log \frac{\pi_{\theta}(y_l \!\mid \!x)}{\pi_{\text{ref}}(y_l \!\mid \!x)} 
\Bigg) \Bigg]
\end{equation}
Pair-DPO simultaneously optimizes understanding and generation using pairedcpreference data, with its loss function comprising these two components:
\begin{equation}
\begin{aligned}
&\mathcal{L}_{\text{Pair-DPO}}(\theta) = \mathcal{L}_{Und}(\theta) + \mathcal{L}_{Gen}(\theta)\\
=&-\!\mathbb{E}_{(x,y, x_w, x_l, y_w, y_l) \sim \mathcal{D}} \!\Bigg[\! \log \sigma \Bigg( 
 \beta \log \frac{\pi_{\theta}(y_w \!\mid\! x)}{\pi_{\text{ref}}(y_w\! \mid \!x)} \!-\! \beta \log \frac{\pi_{\theta}(y_l \!\mid \!x)}{\pi_{\text{ref}}(y_l \!\mid \!x)} \Bigg) +\log \sigma \Bigg( 
 \beta \log \frac{\pi_{\theta}(x_w \!\mid\! y)}{\pi_{\text{ref}}(x_w\! \mid \!y)} \!-\! \beta \log \frac{\pi_{\theta}(x_l \!\mid \!y)}{\pi_{\text{ref}}(x_l \!\mid \!y)} \Bigg) \Bigg]\\
 =&-\!\mathbb{E}_{(x,y, x_w, x_l, y_w, y_l) \sim \mathcal{D}} \!\Bigg[ \!\log \sigma \Bigg( 
 \beta \log \frac{\pi_{\theta}(y_w \!\mid\! x)}{\pi_{\text{ref}}(y_w\! \mid \!x)} \!-\! \beta \log \frac{\pi_{\theta}(y_l \!\mid \!x)}{\pi_{\text{ref}}(y_l \!\mid \!x)} \Bigg) \Bigg( 
 \beta \log \frac{\pi_{\theta}(x_w \!\mid\! y)}{\pi_{\text{ref}}(x_w\! \mid \!y)} \!-\! \beta \log \frac{\pi_{\theta}(x_l \!\mid \!y)}{\pi_{\text{ref}}(x_l \!\mid \!y)} \Bigg) \Bigg]
\end{aligned}
\end{equation}
Here, \(\Delta_{Und}\) and \(\Delta_{Gen}\) are defined as:
\begin{equation}
    \Delta_{Und} = \beta \log \frac{\pi_\theta(y_w \mid x)}{\pi_{\text{ref}}(y_w \mid x)} 
        - 
        \beta \log \frac{\pi_\theta(y_l \mid x)}{\pi_{\text{ref}}(y_l \mid x)} 
\end{equation}
\begin{equation}
    \Delta_{Gen} = \beta \log \frac{\pi_\theta(x_w \mid y)}{\pi_{\text{ref}}(x_w \mid y)} - 
        \beta \log \frac{\pi_\theta(x_l \mid y)}{\pi_{\text{ref}}(x_l \mid y)} 
\end{equation}
Substituting these definitions, the final Pair-DPO objective can be expressed as:
\begin{equation}
    \mathcal{L}_{\text{Pair-DPO}}(\theta ) = -\mathbb{E}_{(x, y, x_w, x_l, y_w, y_l) \sim \mathcal{D}} \left[ 
    \log \sigma \left( 
        \Delta_{Und}\Delta_{Gen}
    \right) 
\right]
\end{equation}
\newpage

\section{Example of Paired Preference Data}
\begin{figure}[h!]
\vskip -0.0in
\begin{center}
\centerline{\includegraphics[width=.87\textwidth]{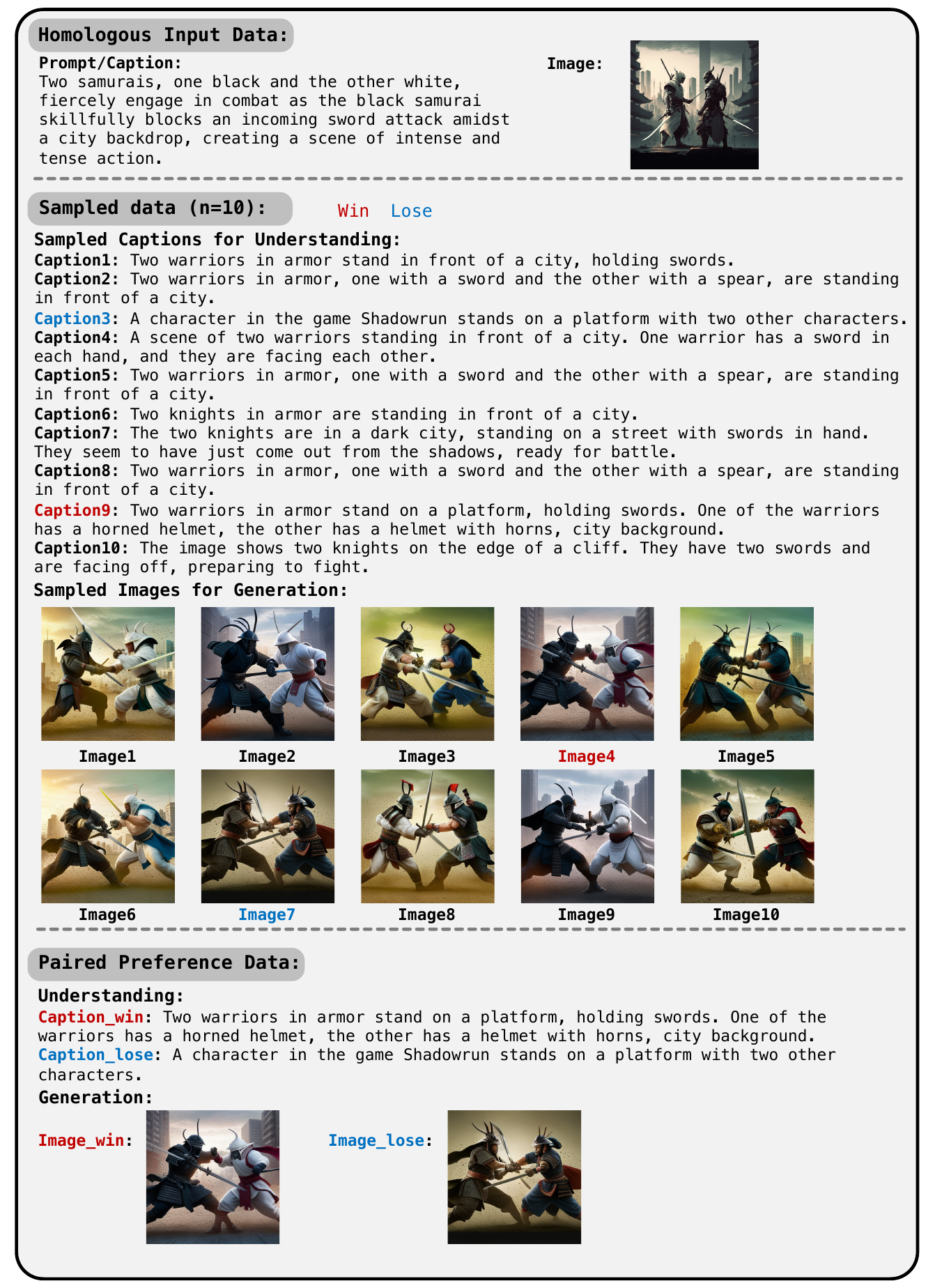}}
\caption{An example of the curation of paired preference data.}
\label{fig:sample}
\end{center}
\vskip -0.3in
\end{figure}

\end{document}